\documentclass[runningheads]{llncs}
\usepackage[T1]{fontenc}
\usepackage{graphicx}
\usepackage{subfig}
\begin{document}
\title{TBPos: Dataset for Large-Scale Precision Visual Localization}
\author{Masud Fahim$^1$, Ilona Söchting$^1$, Luca Ferranti$^1$, Juho Kannala$^2$, Jani Boutellier$^1$}
\authorrunning{M. Fahim, I. Söchting et al.}
\institute{$^1$ University of Vaasa, Vaasa, Finland\\
\{masud.fahim, ilona.sochting, luca.ferranti, jani.boutellier\}@uwasa.fi\\
$^2$ Aalto University, Espoo, Finland\\
juho.kannala@aalto.fi}
\maketitle              % typeset the header of the contribution

\begin{abstract}
Image based localization is a classical computer vision challenge, with several well-known datasets. Generally, datasets consist of a visual 3D database that captures the modeled scenery, as well as query images whose 3D pose is to be discovered. Usually the query images have been acquired with a camera that differs from the imaging hardware used to collect the 3D database; consequently, it is hard to acquire accurate ground truth poses between query images and the 3D database. As the accuracy of visual localization algorithms constantly improves, precise ground truth becomes increasingly important. This paper proposes TBPos, a novel large-scale visual dataset for image based positioning, which provides query images with fully accurate ground truth poses: both the database images and the query images have been derived from the same laser scanner data. In the experimental part of the paper, the proposed dataset is evaluated by means of an image-based localization pipeline.
\end{abstract}

\section{INTRODUCTION}
\label{sec:intro}

Image based localization is one of the enabling technologies for autonomous vehicles \cite{li2018stereo}, augmented reality \cite{cavallari2019real}, and robotics \cite{naseer2018robust}. Driven by advances in deep learning \cite{arandjelovic2016netvlad} \cite{sarlin2019coarse} \cite{sarlin2020superglue} \cite{detone2018superpoint}, the precision of image based localization has progressed in significant leaps during the recent years: for example, using the classical Aachen Day-Night dataset, the 2016 state-of-the-art visual localization approach Active Search \cite{sattler2016efficient} yielded 43.9\% accuracy for night images, whereas HFNet \cite{sarlin2019coarse} proposed in 2019 yielded\footnote{5.0 m distance and 10 degree orientation threshold} already 72.4\% accuracy \cite{sarlin2019coarse}.

\begin{figure} [t]
\centering
\includegraphics[width=\columnwidth]{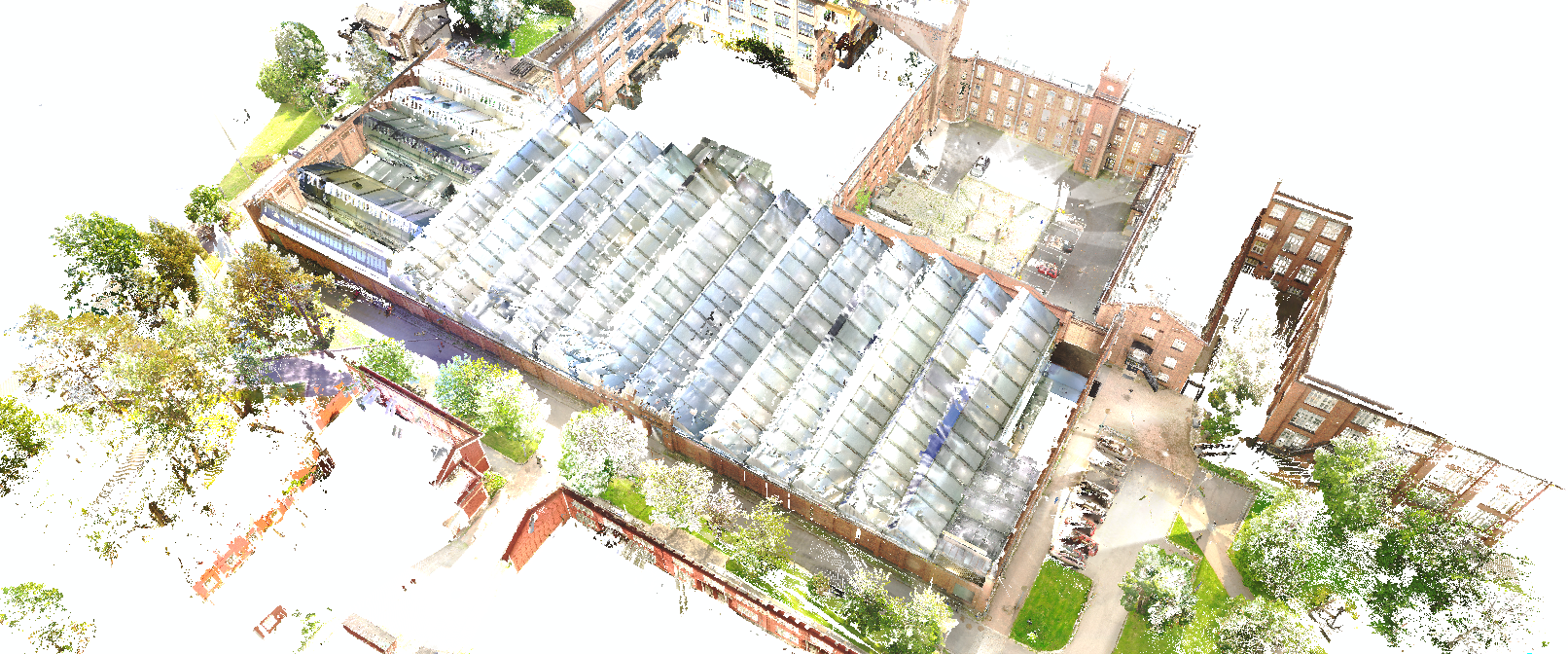}
\caption{Overview of the laser-scanned building, acquired by registering the 182 point clouds that consist the 3D dataset.}
\label{fig:overview}
\end{figure}

To increase the level of challenge for state-of-the-art visual localization algorithms, two main directions of development can be identified: 1) introducing more challenging datasets (for example, by visual data sparsity, occlusions, lighting changes), such as InLoc \cite{taira2018inloc}, and 2) reducing the pose correctness threshold to, e.g., 0.5 m and 2 degrees \cite{dusmanu2019d2}. However, the prerequisite for pose correctness assessment is accurate ground truth (GT); a recent work \cite{brachmann2021limits} pointed out that with most current visual positioning datasets the ground truth poses have been acquired by a reference algorithm such as SfM (e.g., \cite{schonberger2016structure}) or SLAM (e.g., \cite{schops2019bad}). Consequently, the authors of \cite{brachmann2021limits} conclude that benchmarks, which rely on such \textit{pseudo ground truth}, measure how well visual localization algorithms are able to reproduce the output of the reference algorithm (SfM or SLAM), instead of absolute pose accuracy. 

This paper proposes a new large-scale dataset for visual positioning and provides exact ground truth by the use of \textit{synthetic query images}. In the proposed dataset the 3D structure of the environment (see Fig.~\ref{fig:overview}) has been captured using a high-resolution 3D laser scanner, and formatted into \textit{database images} that describe the 3D environment for visual localization methods. Then, instead of relying on secondary camera equipment for capturing \textit{query images}, the query images are generated algorithmically from the same 3D laser scanner data, yielding exact ground truth poses. Based on our experiments, there clearly is a need for exact ground truth: the tested visual localization algorithm \cite{taira2018inloc} was able to reach a localization accuracy of less than 0.02 m and 0.3 degrees, which is more than an order of magnitude less than the presently dominant threshold of 0.25 m and 10 degrees. Besides ground truth accuracy, synthetic query image generation allows producing an arbitrary number of new query images, which can be beneficial for future developments in visual localization.

Obviously, the proposed approach of using synthetic images risks creating either a) too easy or b) unrealistic query images. However, our claim is that the generated query images are sufficiently realistic, because they have been formed from measured visual data, and their degree of challenge can be increased by introducing a set of irreversible distortions, such as occlusion or nonlinear lighting variation. In the experimental section of the paper we apply the InLoc \cite{taira2018inloc} visual localization pipeline to the proposed dataset to assess its level of challenge compared to the well-known InLoc dataset \cite{taira2018inloc}.

\begin{table*} [t]
\begin{center}
\caption{Statistics of selected large-scale localization datasets: number of locations used to acquire database images, number of provided database and query images, means for acquiring ground truth (GT) poses; P3P: Perspective-three-point \cite{gao2003complete}, BA: Bundle adjustment, SfM: Structure-from-Motion \cite{schonberger2016structure}, VS: View synthesis \cite{taira2018inloc}, ICP: Iterative closest point \cite{besl1992method}, SIFT: Scale-invariant feature transform \cite{lowe2004distinctive}. (*) The number of query images synthesized for experiments presented in this work.}
\label{table:datasets}
\begin{tabular}{p{4.8cm}p{1.8cm}p{1.8cm}p{1.6cm}p{1.6cm}}
\hline\noalign{\smallskip}
Dataset & \#Locations & \#Database & \#Query & GT\\
\noalign{\smallskip}
\hline
InLoc \cite{taira2018inloc} & 277 & 9972 & 329 & P3P+BA\\
Aachen Day-Night \cite{sattler2012image} \cite{sattler2018benchmarking} & n/a & 4328 & 922 & SfM+P3P\\
Aachen Day-Night v1.1 \cite{sattler2012image} \cite{zhang2021reference} & n/a & 6697 & 1015 & SfM+VS\\
RobotCar Seasons \cite{maddern20171} \cite{sattler2018benchmarking} & 8707 & 26121 & 11934 & ICP\\
CMU Seasons \cite{badino2011visual} \cite{sattler2018benchmarking} & 17 & 7159 & 75335 & SIFT+BA\\
\hline
TBPos (proposed) & 182 & 6552 & 338* & VS\\
\hline
\end{tabular}
\end{center}
\end{table*}

The contributions of this work are as follows:
\begin{enumerate}
    \itemsep0em
    \item TBPos, a novel open\footnote{
    https://gitlab.com/jboutell/tbpos; https://doi.org/10.5281/zenodo.7466448} 3D dataset for large-scale visual precision localization,     
    \item Proposed approach of view synthesis for query image and exact ground truth generation, and
    \item Benchmarking of the proposed dataset using a visual localization pipeline.
\end{enumerate}

Furthermore, our results point out that the use of view synthesis enables generating challenging query images from viewpoints, where traditional query image acquisition could not provide reliable ground truth. For the development of practical image based localization applications (e.g., autonomous vehicles) such challenging real-life cases need to be considered as well.

\section{Related Work}
\label{sec:related}

\subsection{Datasets}

Several well-known datasets have been published for large-scale visual localization, some of which are shown in Table~\ref{table:datasets}. Related to this work, the most significant differences between the datasets can be seen in the means how ground truth poses for query images have been acquired. For the RobotCar Seasons dataset \cite{maddern20171} \cite{sattler2018benchmarking}, query image ground truth poses were acquired from LIDAR-based point clouds after ICP \cite{besl1992method} alignment and some manual adjustment \cite{sattler2018benchmarking}. For the CMU Seasons dataset \cite{sattler2018benchmarking}, bundle-adjusted SIFT \cite{lowe2004distinctive} features were used to construct local 3D scenery models, and to acquire GT poses. In the InLoc dataset \cite{taira2018inloc}, GT poses were acquired using the P3P-RANSAC and bundle adjustment, with manual matching for difficult cases. The Aachen Day-Night dataset \cite{sattler2018benchmarking} exists in two versions, for both of which GT poses were estimated from an SfM model using a P3P solver. The recent v1.1 version \cite{zhang2021reference} of the Aachen Day-Night dataset comes closest to the proposed work in the sense that it uses view synthesis \cite{taira2018inloc}; however, whereas \cite{zhang2021reference} uses view synthesis for refinement of GT poses related to separately acquired query images, in our proposed dataset view synthesis is used to \textit{render the query images}. Whereas the approach taken in \cite{zhang2021reference} enables more accurate reference pose acquisition for manually acquired query images, the proposed approach enables automatic generation of query images with exact ground truth. A further advantage of the proposed approach is its applicability to the training of machine learning based visual localization  techniques that require numerous query images, such as \cite{ferranti2021can}.

\subsection{Algorithms}

Algorithms for image based localization can be classified into \textit{direct matching}, \textit{retrieval-based} and \textit{learning-based} approaches \cite{ferranti2019confidence}.

The InLoc \cite{taira2018inloc} work presented both an extensive dataset for visual localization, and an image retrieval-based localization pipeline.
In retrieval based pipelines, the main stages are \textit{feature extraction}, \textit{image retrieval}, \textit{dense matching} and \textit{pose estimation} \cite{ferranti2021can}. Whereas the earlier pipeline stages between feature extraction and dense matching are based on learned image features, the pose estimation step leverages classical geometric computer vision for estimating the 6DoF camera pose for a given query image. InLoc appended this generic pipeline with the \textit{pose verification} step that further increases localization accuracy. Recently, the work PCLoc \cite{hyeon2021pose}, built on InLoc, also added a further \textit{pose correction} step between the stages of pose estimation and pose verification.

The learning-based work HFNet \cite{sarlin2019coarse} proposed a monolithic CNN for simultaneous keypoint detection, as well as local and global keypoint extraction, which provides significant runtime computation savings. Despite computational efficiency, results reported in \cite{hyeon2021pose} show that on average HFNet yields a similar level of accuracy as the InLoc \cite{taira2018inloc} visual localization algorithm.

\begin{figure} [t]
\centering
\includegraphics[width=\columnwidth]{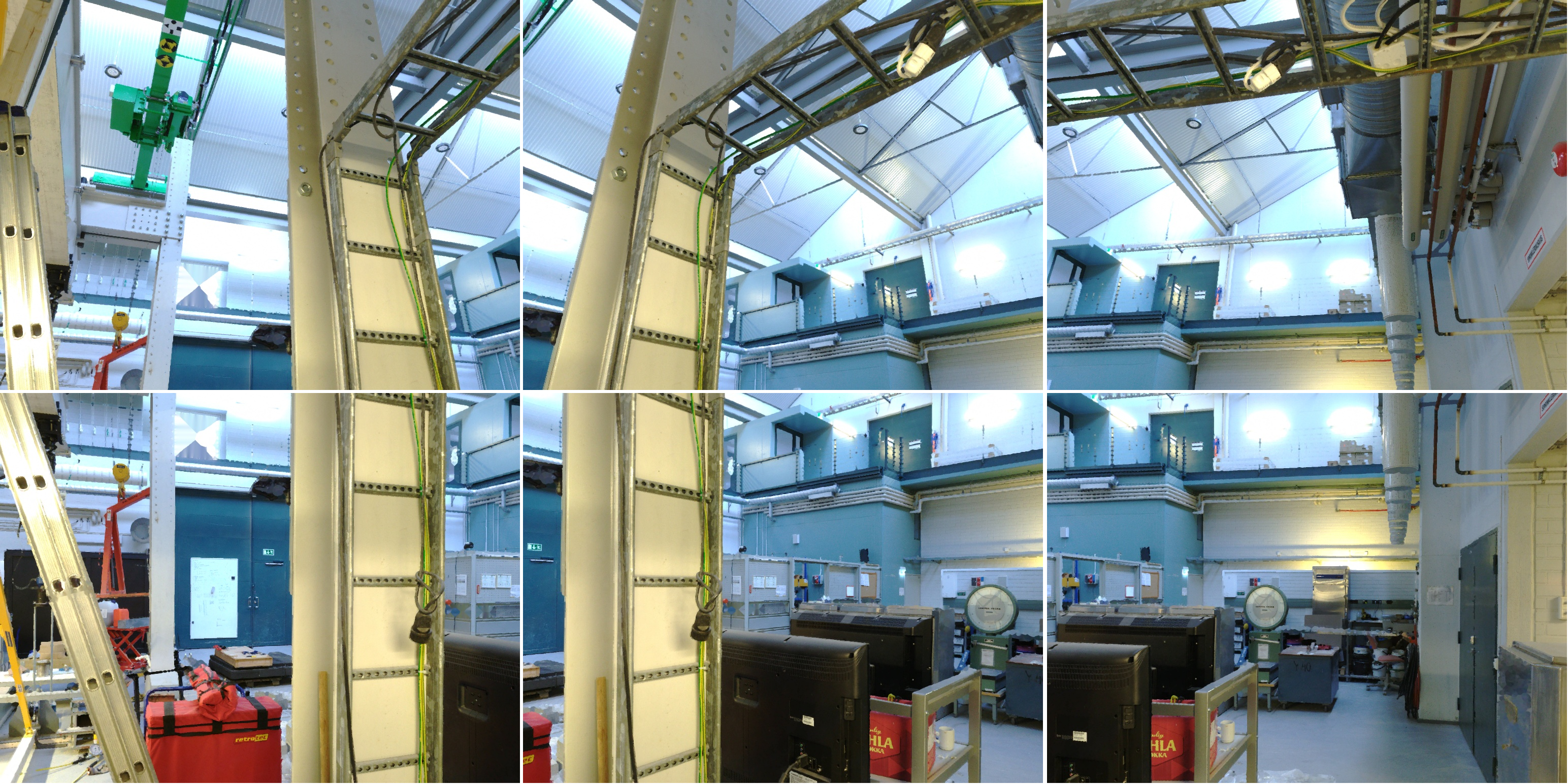}
\caption{Six database images extracted from a single point cloud scan. Each image has a $60^{\circ}$ field-of-view (hor.), and the images have been taken with a $30^{\circ}$ stride.}
\label{fig:cutouts}
\end{figure}

\section{Visual Data Acquisition Procedure}
\label{sec:dataset}

The visual data for TBPos was acquired by laser-scanning an industrial building (see Fig.~\ref{fig:overview}), which consists of inside and outside areas. Typical to such buildings, both repeating and absent texture are commonplace on surfaces. The 3D structure contains a lot of recurring shapes: identically-shaped rooms, tie beams, etc.
Compared to the related datasets of Table~\ref{table:datasets}, the proposed TBPos dataset is characteristically most similar to the InLoc dataset, with the most significant difference that TBPos covers also outside areas.

The laser scanning was performed using a Faro Focus 3D scanner, providing colored point clouds that mostly contain 27 million points each. Individual point clouds were registered against each other using the Faro SCENE software that required some degree of manual correspondence annotation.

To ease the adoption of TBPos, the data structure of InLoc \cite{taira2018inloc} was used for point clouds, database images, query images and supporting data. For database images, each laser scan was sliced into 36 perspective RGBD images with $30^{\circ}$ sampling stride and $60^{\circ}$ field-of-view. The chosen resolution for database images was 1024$\times$768. An example of six consecutive database images is shown in Fig.~\ref{fig:cutouts}.

\begin{figure} [t]
\centering
\includegraphics[width=\columnwidth]{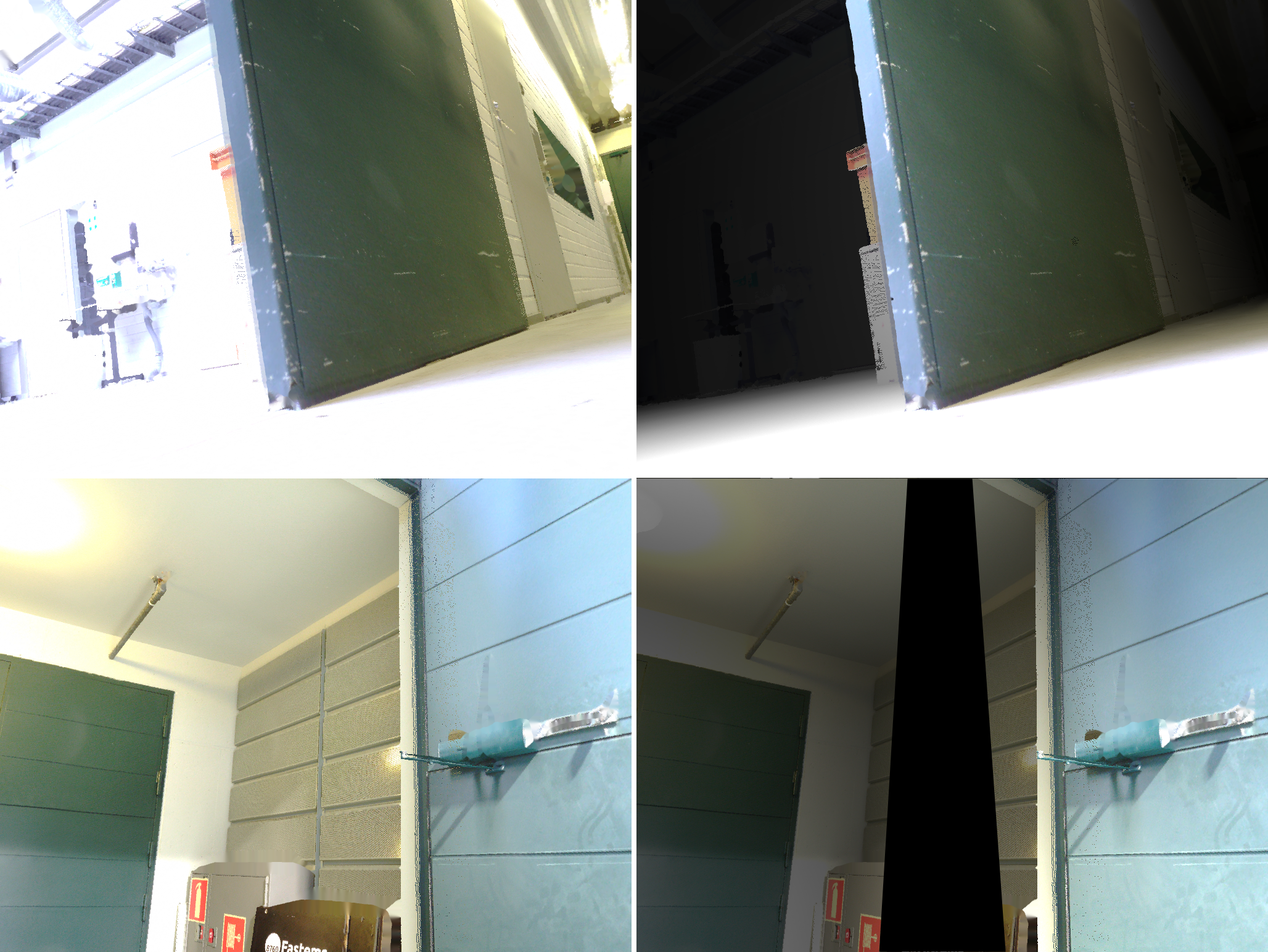}
\caption{Synthesized query image samples. Left-hand images have been extracted from the laser scanner data without modification, whereas their right-hand counterparts are after lighting adjustment and addition of 2D occlusion (lower-right image).}
\label{fig:views}
\end{figure}

\begin{figure} [t]
\centering
\includegraphics[width=\columnwidth]{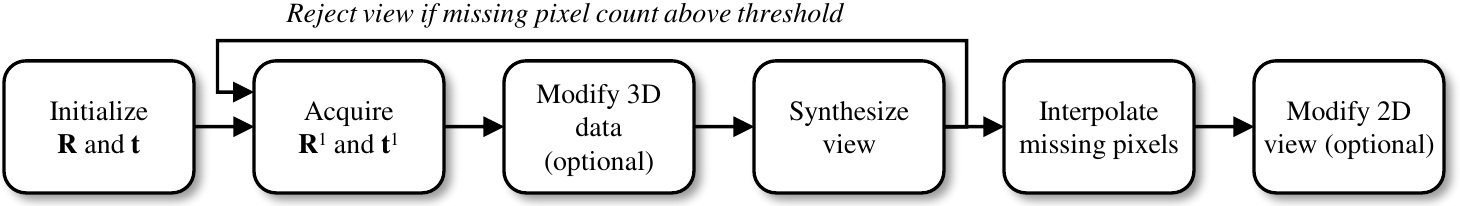}
\caption{The proposed procedure for query image synthesis.}
\label{fig:pipeline}
\end{figure}

\section{Synthesizing Query Images}
\label{sec:synthesis}

The adopted procedure of query image synthesis (see Fig.~\ref{fig:pipeline}) is similar to the procedure of database image generation. Starting from the original laser scanner position $\textbf{t}$, the virtual camera position is randomly perturbed along X, Y and Z axes, providing a new position $\textbf{t}^1$. Similarly, starting from initial laser scanner pointing direction
$\textbf{R}$, a random view direction $\textbf{R}^1$ is generated, forming the virtual camera 6DOF pose $\textbf{R}^1\textbf{t}^1$ \textit{that we also adopt as the ground truth for pose estimation benchmarking}. After acquiring $\textbf{R}^1 \textbf{t}^1$, the 3D environment can be modified for the purpose of generating query images such that they reflect challenges relevant to practical pose estimation, e.g., lighting variation, displaced objects, etc. 

Provided $\textbf{R}^1 \textbf{t}^1$, a predefined focal length \textit{f}, and the desired query image resolution $\{ r_x, r_y \}$, a novel view can be synthesized from the surrounding 3D point cloud. Due to various reasons, such as occluded view directions, a certain proportion of synthetic view pixels can remain absent of visual data. In our proposed view synthesis approach these missing pixels are filled using an iterative, clamping based interpolation procedure. However, in order to maintain high visual quality for the synthesized query images, randomly generated virtual camera orientations $\textbf{R}^1 \textbf{t}^1$ that have too much missing visual information can be discarded by straightforward missing pixel counting. As an optional last step, the generated 2D view can be further modified to increase the positioning challenge. Such modification can include occlusion, lighting variation or image noise.

In the practical implementation of the proposed view synthesis approach, the random view rotations $\textbf{R}^1$ around the Y (horizontal) axis were restricted in order to avoid generating views that dominantly show floor or ceiling. Similarly, perturbation of camera location was limited to small translations of a couple of meters to avoid placing the virtual camera behind walls.

\section{Experimental Results}
\label{sec:experiments}

For benchmarking purposes a series of 338 query image poses were generated using the procedure described in Section~\ref{sec:synthesis}. The query images were chosen to be sampled with a view angle and resolution identical to the database images ($60^{\circ}$, 1024$\times$768), although there was no technical restriction to select other values. In order to increase the pose estimation challenge, the data was modified both in 3D (point cloud format) and 2D in the query image synthesis process.

The 3D point cloud data was lighting-adjusted to simulate a situation where an autonomous mobile platform navigates in a dark environment, lighting its surroundings by in-built lights. In Fig.~\ref{fig:views}, the two topmost images show the effect of this operation: brightness of 3D points decreases as a function of camera-point distance, simulating the effect of a flashlight close to the camera. In terms of pose estimation challenge, this effect causes the appearance of distant query image features to change considerably compared to their database counterparts. On the other hand, lightly-colored surfaces close to the camera tend to get overexposed, losing similarity with the respective database images.

Besides 3D lighting adjustment, also additional occlusions were introduced to most query images: an example of this can be seen in the lower-right image of Fig.~\ref{fig:views}. Each query image to be occluded was allowed to get a random quadrangle to cover between 1\% and 50\% of query image area, simulating the case where an object gets in between the surroundings and the autonomous platform's camera.

A significant benefit in using synthetic query images instead of manually acquired ones can be discovered by comparing the TBPos queries to the queries of the InLoc dataset \cite{taira2018inloc}: whereas InLoc queries have exclusively been taken at human eye level with unnoticeable changes in camera pitch or roll, the TBPos query image locations range from close-to-floor level (Fig.~\ref{fig:views}) to higher altitudes, and include random rotations along all camera axes.

To assess the difficulty level of the query image set, the InLoc visual localization pipeline was used to acquire localization accuracy at the commonly applied translation thresholds of 0.25 m, 0.5 m and 1.0 m, and angular threshold of 10$^{\circ}$. The results are shown in Table~\ref{table:accuracy}(a) together with the respective results for the InLoc dataset. The numbers show that the proposed TBPos dataset with the 338 query images are significantly more challenging than the InLoc dataset, as localization success at 1.0 m and $10^{\circ}$ is more than 30\% lower for TBPos.

Table~\ref{table:accuracy}(b) details the TBPos pose estimation accuracy analysis at different stages of the InLoc pose estimation pipeline. It can be seen that the impact of the pose verification (PV) stage is minor, ranging between 0.9\% and 2.4\% in the success rate of $\le 1.0$ m accuracy. The 'Top10' column of Table~\ref{table:accuracy}(b), on the other hand, shows how often at least one database image \textit{from the same point cloud scan as the query image}, appears in the list of top-10 database image candidates. For example, 66.6\% means that for 225 out of the 338 query images, the InLoc pipeline has ranked at least one database image from the same point cloud scan as the query, into top-10 best candidates, and consequently this figure measures the success rate of image retrieval (IR).

With the success rate of 63.9\%, the InLoc IR clearly has some challenges with the TBPos dataset. However, when the actual camera pose is estimated by dense matching (DM) for the best matching database image candidate using one of the metric thresholds, the success rate drops by around 30\%. Consequently, it can be stated that for the TBPos set of 338 query images, the pose estimation difficulties are related both to IR and DM stages.

The precision of the InLoc pose estimation pipeline can be evaluated by computing the average deviation from ground truth for the 29.6\% of cases where the query image is within 0.25 m and 10$^{\circ}$ from ground truth: for these success cases the average location deviation is 0.10 m and the average angular deviation is 2.26$^{\circ}$ from the ground truth for the TBPos set of 338 query images. In order to get better understanding of visual localization accuracy potential, an easier version of the 338 query image set was generated using the same ground truth poses, but without lighting variation or occlusions. With this considerably easier set of query images, the top-30\% of pose estimates reached a localization accuracy below 0.02 m and angular accuracy below 0.3 degrees. Evidently, accuracy analysis like this requires exact ground truth that is available by the proposed approach of query image synthesis. 

Still, also with synthesized query images, the achievable lower bound of accuracy is determined by the precision of the imaging device used for dataset collection, as well as the resolution of the database and query images.

\begin{table}[t]
  \centering
  \subfloat[][InLoc localization pipeline accuracy at various thresholds for the proposed and reference (InLoc) datasets using the $10^{\circ}$ angular threshold.]{
\begin{tabular}{p{1.4cm}p{0.9cm}p{0.9cm}p{0.9cm}}
\hline\noalign{\smallskip}
Dataset & 0.25m & 0.5m & 1.0m \\
\noalign{\smallskip}
\hline
InLoc \cite{taira2018inloc} & 41.6\% & 56.5\% & 67.2\% \\
\hline
TBPos & 29.6\% & 34.3\% & 35.2\% \\
\hline
\end{tabular}  
  }%
  \qquad
  \subfloat[][InLoc pose estimation success rate at different pipeline stages, for the TBPos dataset. IR: image retrieval, DM: dense matching, PV: pose verification.]{
\begin{tabular}{p{2.0cm}p{0.9cm}p{0.9cm}p{0.9cm}p{0.9cm}}
\hline\noalign{\smallskip}
TBPos datas.& 0.25m & 0.5m & 1.0m & Top10\\
\noalign{\smallskip}
\hline
IR+DM+PV & 29.6\% & 34.3\% & 35.2\% & 66.6\%\\
\hline
IR+DM & 27.2\% & 33.4\% & 34.3\% & 66.6\%\\
\hline
IR & n/a & n/a & n/a & 63.9\%\\
\hline
\end{tabular}
}
\caption{TBPos evaluation.}
\label{table:accuracy}
\end{table}

\begin{figure} [t]
\centering
\includegraphics[width=\columnwidth]{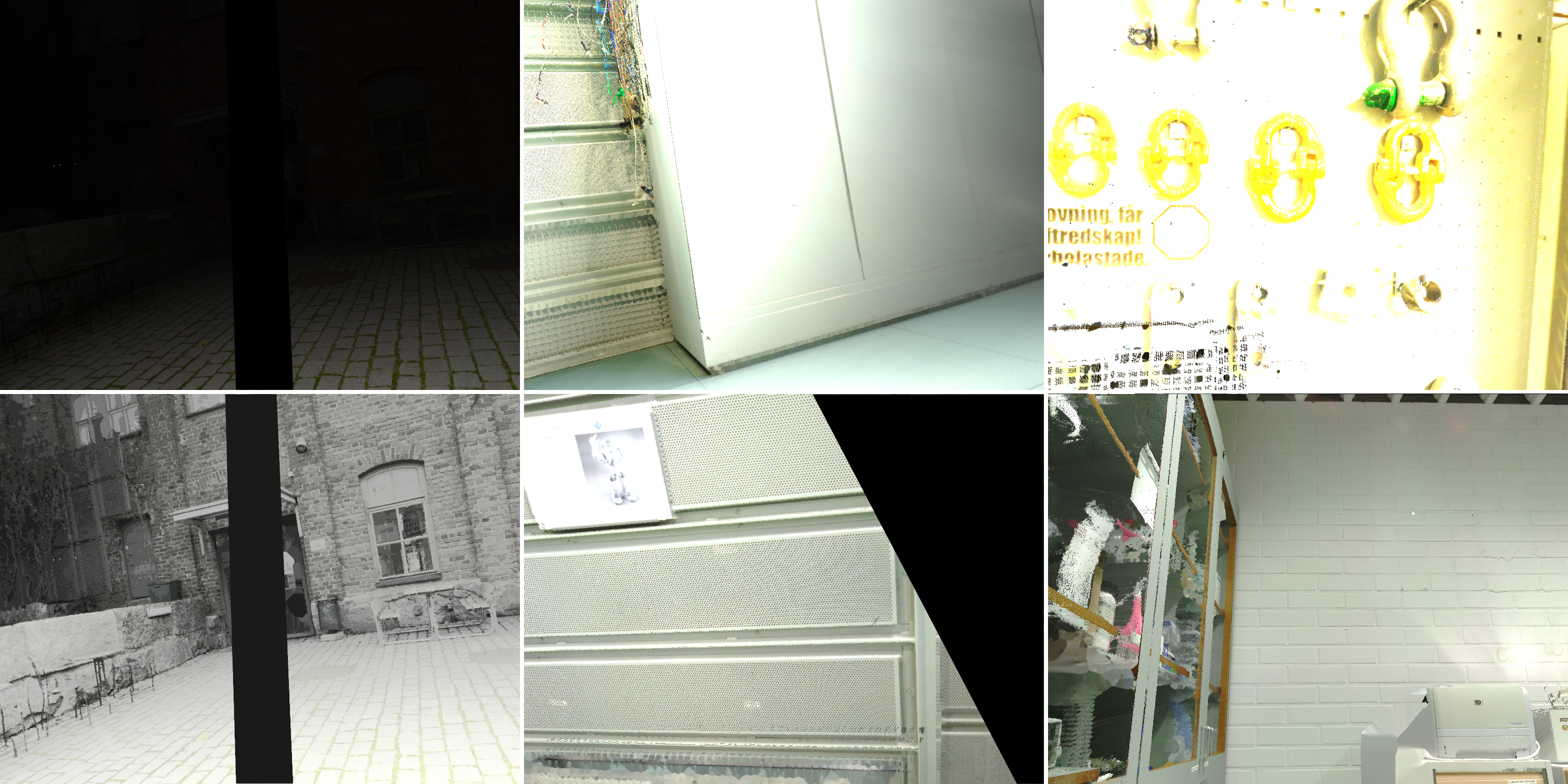}
\caption{Examples of TBPos query images where the InLoc pose estimation pipeline completely fails.}
\label{fig:pv}
\end{figure}

\subsection{Analysis of Pose Estimation Failure Cases}

Fig.~\ref{fig:pv} depicts representative samples from the set of 338 queries, where pose estimation by InLoc failed. The top-left image shows a query from outside environment, where distance-based darkening has rendered distant visual details almost invisible. For this query image, InLoc is not capable of computing any pose. However, if the brightness of the query image is manually increased using regular image processing (see Fig.~\ref{fig:pv}, bottom-left), ample visual detail is revealed and manual localization would be possible. Hence, we consider the pose estimation failure of this case a shortcoming of the InLoc pose estimation pipeline. 

The top-center and bottom-center cases of Fig.~\ref{fig:pv} show two further cases where the InLoc pipeline completely fails in pose estimation. The top-center image shows a large area without any texture, whereas the bottom case exhibits patterns that are repeated in several locations of the whole dataset, and successful localization would require paying attention to small details that discriminate the particular location from similar places elsewhere in the database.

The top-right case shows a close-up query image with very specific detail that would make manual pose estimation straightforward, but also in this case InLoc completely fails  due to the overexposure that makes the query image's features differ significantly from the ones of the respective database image.

Finally, the failure case depicted in the bottom-right corner is particularly interesting, as it contains a high amount of distinct visual detail, yet the pose estimation pipeline fails here as well. Unlike the previous examples, this last example is without darkening or occlusion.

\section{Conclusions}
\label{sec:conclusions}

This paper proposes a novel open dataset, TBPos, for image based large-scale precision localization. In order to achieve exact ground truth for localization algorithm benchmarking, we have adopted the approach of query image synthesis. In the experimental section the proposed dataset has been benchmarked using the InLoc localization pipeline, and has been compared in terms of difficulty to the well-known InLoc dataset, showing that TBPos is significantly more challenging. In addition to measuring the conventional localization success rate, our approach also enables measuring the metric precision of image-base localization.

\par\vfill\par

\clearpage
%\bibliographystyle{splncs04}
%\bibliography{egbib}

\end{document}